\documentclass[11pt,a4paper]{article}
\usepackage[hyperref]{acl2018}
\usepackage{times}
\usepackage{latexsym}

\usepackage{url}

\usepackage{tabularx}

\usepackage{amsmath}
\usepackage{booktabs}
\usepackage{graphicx}
\usepackage{url}
\usepackage[utf8]{inputenc}
\usepackage{wrapfig,lipsum}
\usepackage{graphicx}
\usepackage{pgfplots}
\usetikzlibrary{patterns}
\usepackage{subcaption}
\usepackage{textcomp}

\aclfinalcopy 

\title{Translating Questions into Answers using DBPedia n-triples}

\author{Mihael Arcan \\
  Insight Centre for Data Analytics \\
  National University of Ireland Galway \\
  {\tt mihael.arcan@insight-centre.org} }

\date{}

\begin{document}
\maketitle

\begin{abstract}
In this paper we present a question answering system using a neural network to interpret questions learned from the DBpedia repository. We train a sequence-to-sequence neural network model with n-triples extracted from the DBpedia Infobox Properties. Since these properties do not represent the natural language, we further used question-answer dialogues from movie subtitles. Although the automatic evaluation shows a low overlap of the generated answers compared to the gold standard set, a manual inspection of the showed promising outcomes from the experiment for further work.
\end{abstract}

\maketitle

\section{Introduction}
\label{sec:intro}

Recent work on sequence-to-sequence neural networks have shown remarkable progress in many domains such as speech recognition,  computer  vision  and  language  processing. These neural models can be used to map a sequence to another sequence, which has direct applications in natural language understanding \cite{Sutskever:2014:SSL:2969033.2969173}. One of the major advantages of this framework is that it requires a limited amount of feature engineering while still improving state-of-the-art results. This allows researchers to work on tasks for which domain knowledge may not be available. Question answering (QA) systems can directly benefit from this novel approaches, because it only requires mapping between questions and their answers. Due to the complexity of this mapping, most of the previous research in this domain used a complex pipeline of conventional linguistically-based NLP techniques, such as parsing, part-of-speech tagging and co-reference resolution \cite{Unger2014}.

In this work, we build an open-domain QA system with sequence-to-sequence neural network models. The system was trained with question-answer pairs on the world knowledge represented in the DBpedia repository. Although building an open-domain system allows us to use an extensive amount of data stored in DBpedia, one of the challenges is to store this large amount of knowledge in the neural network architecture. Although these systems generally involve a smaller learning pipeline, they require a significant amount of training data.

\begin{figure}
 \centering
 \includegraphics[width=0.5\textwidth]{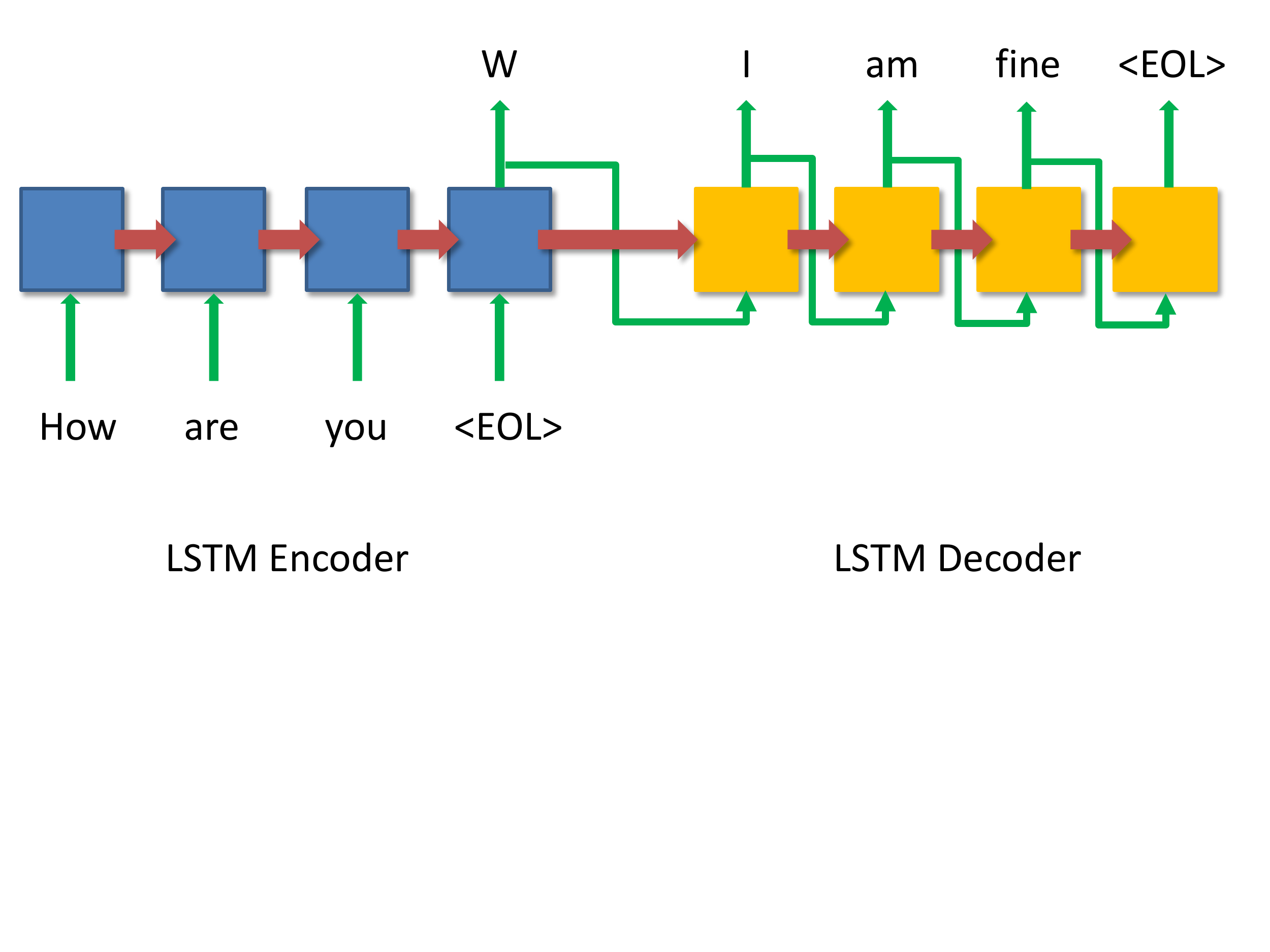}
 \vspace{-30mm}
 \caption{Neural network with the encoder-decoder architecture.}
 \vspace{-5mm}
 \label{fig:graph}
\end{figure}

Figure~\ref{fig:graph} illustrates how a sequence-to-sequence neural network can be trained on question-answering  pairs. First, an sequence-to-sequence framework reads the source sentence, i.e. question, using an encoder to build a dense vector, a sequence of non-zero values that represents the meaning of a question. A decoder, processes this vector to predict an answer. In this manner, these encoder-decoder models can capture long-range dependencies in languages, e.g., gender agreements or syntax structures.

\section{Related Work}
\label{sec:relwork}
Recent approaches on building QA systems are dominated by the usage of neural networks. \cite{journals/corr/VinyalsL15} present an approach for conversational modelling, which uses a sequence-to-sequence neural model. Their model predicts the next sentence given the previous sentences for an IT helpdesk domain, as well as for an open-domain trained on a subtitles dataset. For an open-domain dialogue generation, \citep{DBLP:conf/emnlp/LiMSJRJ17} propose using adversarial training. Therefore, reinforcement learning is used to train the system that produces sequences that are indistinguishable from human-generated dialogue utterances. For this, they jointly train two systems, a generative model, which produces response sequences and a discriminator to distinguish between the human-generated dialogues and the machine-generated ones. The outputs from the discriminator are then used as rewards for the generative model, guiding the system to generate dialogues that mostly resemble human dialogues. \cite{Yin:2016:NGQ:3060832.3061037} demonstrate an end-to-end neural network model for generative QA. Their model is built on the encoder-decoder framework for sequence-to-sequence learning, while equipped with the ability to query a knowledge-base, which they demonstrate on the Chinese encyclopedia web site. The authors show that the proposed model is capable of generating natural and right answers by referring to the facts in the knowledge base. Additionally to the previous approaches, \cite{Serban:2016:BED:3016387.3016435} extend the hierarchical recurrent encoder-decoder neural network to the open domain dialogue system and demonstrate that this model is competitive with state-of-the-art neural language models and back-off n-gram models. They illustrate limitations of similar approaches and show how the performance can be improved by bootstrapping the learning from a larger question-answer pair corpus and from pre-trained word embeddings.
A heuristic that guides the development of neural baseline systems for the extractive QA task is described in \cite{Weissenborn2017MakingNQ}, which serves as guideline for the development of two neural baseline systems. Their RNN-based system, called FastQA, demonstrates good performance for extractive question answering due to the awareness of question words while processing the context. Additionally they introduce a composition function that goes beyond simple bag-of-words modelling. \cite{W17-6935} demonstrate an approach to non-factoid answer generation with a separate component, which bases on BiLSTM to determine the importance of segments in the input. In contrast to other attention-based models, they determine the importance while assuming the independence of questions and candidate answers. \cite{P17-1168} present their Gated-Attention reader for answering cloze-style questions over documents. The reader features a novel multiplicative gating mechanism in combination with a multi-hop architecture, which is based on multiplicative interactions between the query embedding and the intermediate states of a recurrent neural network document reader. This enables the reader to build query-specific representations of tokens in the document for accurate answer selection. 

The work mentioned above employ neural networks, but incorporate only a small amount of question-answering pairs to build their QA systems. Work on QA, which uses the DBpedia knowledge on the other hand, use much more complex linguistically-based NLP techniques to generate an answer for a given question. \cite{7845035} propose a general purpose a QA system using Wikipedia data as its knowledge source, which can answer wh-interrogated questions. Their QA system includes tasks of named entity tagging, question classification, information retrieval and answer extraction. 
\cite{7600312} present a QA approach on the DBpedia dataset. With predefined templates and dependency parsing on questions they obtain the entity and property mention from the question. They split the QA process into two steps; first they build an entity-centric index to search the target entity, whereby in the second step they expand the property mention with WordNet and ConceptNet to match the target entity with the searched entity property. Similarly, \cite{DBLP:conf/clef/2013w} present a QA system over Linked Data (DBpedia), which focuses on construct a bridge between the users and the Linked Data. Based on the consisting of subject-property-object (SPO) triples, each natural language question firstly is transformed into a triple-based representation (query triple). Then, the corresponding resources in DBpedia, including class, entity, property, are mapped for the phrases in the query triples. Finally, the optimal SPARQL query is generated as the output result. With this approach, their system can not only deal with the single-relation questions but also complex questions containing multi-relations. 

Differentially to the approaches mentioned above, we combine the neural network approaches with the large number of facts (> 20M n-triples) extracted from the DBpedia repository.

\section{Experimental Setting}
\label{sec:expsetting}
In this section, we give an overview of the sequence-to-sequence framework used in this experiment as well as information on the datasets used in our experiment. Furthermore, we give insights into the techniques to evaluate the correctness of the provided answers.

\subsection{Sequence-to-sequence Neural Network Toolkit}
\textbf{OpenNMT}~\cite{2017opennmt} is a generic deep learning framework mainly specialised in sequence-to-sequence (seq2seq) models covering a variety of tasks such as machine translation, summarisation, speech processing and question answering. We used the default neural network training parameters, i.e. 2 hidden layers, 500 hidden LSTM units, input feeding enabled, batch size of 64, 0.3 dropout probability and a dynamic learning rate decay. 
We train the network for 13 epochs and report the results in Section \ref{sec:eval}.

\paragraph{\textbf{Data Compression - Byte Pair Encoding}}
A common problem in training a neural network is the computational complexity, which cause that the vocabulary has to be limited to a specific threshold. Because of this the neural network can not learn expressions of rare and unknown words, e.g. domain-specific expressions. Therefore, if the training method does not see a specific word or phrase multiple times during training, it will not learn the interpretation of the word. This challenge is even more evident in sequence-to-sequence models used for summarisation, question answering or machine translation. Therefore the vocabulary is often limited only to 50,000 or 100,000 words (in comparison to 300,000 or more unique words in our training data sets, see Table~\ref{tab:data_stats}). To overcome this limitation, different methods were suggested, i.e. character based neural model \cite{DBLP:journals/corr/Costa-JussaF16,DBLP:journals/corr/LingTDB15} or using subword units, e.g. Byte Pair Encoding (BPE). The latter one was successfully adapted for word segmentation specifically for the NMT scenario \cite{journals/corr/SennrichHB15}. BPE \cite{Gage:1994:NAD:177910.177914} is a form of data compression that iteratively replaces the most frequent pair of bytes in a sequence with a single, unused byte. Instead of merging frequent pairs of bytes as shown in the original algorithm, characters or character sequences are merged for the purposes of natural language generation. To achieve this, the symbol vocabulary is initialised with the character vocabulary, and each word is represented as a sequence of characters—plus a special end-of-word symbol, which allows to restore the original tokenisation after the generating the answer based on the given question. This process is repeated as many times as new symbols are created.

\subsection{Training Datasets for the QA system}
For our QA system we used the DBpedia repository and the Subtitles corpus. Due to the nature of training a sequence-to-sequence neural model, questions and answers need to be aligned. The statistics on the used data are shown in Table~\ref{tab:data_stats}.

\begin{table}
\centering
\small
\setlength{\tabcolsep}{3pt}
\begin{tabular}{c|rcc}
\toprule
DBpedia  & lines & 21,281,382\\
\cmidrule{3-4}
Properties & & Vocabulary & Avg. Words/Line\\
& Questions & 1,511,062 & 6.02 \\
& Answers & 3,189,487 & 3.61\\
\midrule
OpenSubtitles & lines & 5,000,000\\
\cmidrule{3-4}
& & Vocabulary & Avg. Words/Line\\
& Questions & 345,488 & 5.45 \\
& Answers & 490,396 & 5.61\\
\midrule
Validation Set & lines & 20,000\\
\cmidrule{3-4}
& & Vocabulary & Avg. Words/Line\\
& Questions & 21,861 & 5.66\\
& Answers & 22,209 & 4.55\\
\midrule
Evaluation Set & lines & 1,000\\
\cmidrule{3-4}
& & Vocabulary & Avg. Words/Line\\
& Questions & 15,084 & 5.95\\
& Answers & 13,144 & 3.60\\
\bottomrule
\end{tabular}
\caption{Statistics on the datasets used to train, validate and evaluate the sequence-to-sequence models.}
\label{tab:data_stats}
\end{table}

\paragraph{\textbf{DBPedia repository}}
In our work, we use the DBpedia \cite{dbpedia-swj} repository (version 2016-04). The DBpedia project aims to extract structured knowledge from the knowledge added to the Wikipedia repository. DBpedia allows users to semantically query relationships and properties of Wikipedia resources, including links to other related datasets. 

For our experiment on training a QA system, we extracted (see Section~\ref{sec:meth}) the knowledge stored in \textit{Dbpedia Infobox Properties},\footnote{\url{http://downloads.dbpedia.org/2016-04/core-i18n/en/infobox_properties_en.ttl.bz2}} which with its n-triple structure represents answers to particular questions. From this dataset we extracted more than 21M question-answer entries.

\paragraph{\textbf{OpenSubtitles}}
Since the extracted information form the DBpedia n-triples does not represent a natural language, we added to the extracted n-triples answers with dialogues from the OpenSubtitles dataset \cite{Tiedemann:RANLP5}. Since this dataset is stored in an XML structure with time codes, only sentences were extracted, where the first sentence ends with a question mark and the second sentence does not end with a question mark. Additionally, to ensure a better consistency between an question and the answer, the second sentence has to follow the first sentence by less than 20 seconds. From the 14M sentence corpus\footnote{\url{https://s3.amazonaws.com/opennmt-trainingdata/opensub_qa_en.tgz}} of question-answer pairs provided by the OpenNMT project, we used 5M dialogue entries to modify the language generation part. Table~\ref{tab:subtit_qa_examples} shows some examples from the OpenSubtitles dataset.

\begin{table}
\centering
\small
\setlength{\tabcolsep}{2pt}
\begin{tabular}{ll}
\toprule
Question & Answer \\
\midrule
Where's Lane going? & Away. \\
Can you just let me out, man? & I tell you what. \\
You trying to get high? & No. \\
you want to become a priest? & Yeah. \\
So you believe me? & I don't know what to believe. \\
\bottomrule
\end{tabular}
\caption{Examples of the OpenSubtitles question-answering sentence pairs.}
\label{tab:subtit_qa_examples}
\end{table}


\subsection{Evaluation Dataset}
For the automatic evaluation of the QA system, we used 10,000 entries extracted from the DBpedia Infobox Property repository (Table~\ref{tab:data_stats}).

\subsection{Evaluation Metrics}
The automatic evaluation of the proposed QA system is based on the correspondence between the generated answer and gold standard. For the automatic evaluation we used the BLEU~\cite{Papineni:2002}, METEOR \cite{denkowski:lavie:meteor-wmt:2014} and chrF~\cite{popovic:2015:WMT} metrics. \textbf{BLEU} (Bilingual Evaluation Understudy) is calculated for individual generated segments (n-grams) by comparing them with a dataset of of the gold standard. Considering the shortness of the questions, we report besides the four-gram overlap (BLEU) also scores based on the unigram overlap (BLEU-1). Those scores, between 0 and 100 (perfect overlap), are then averaged over the whole \textit{evaluation dataset} to reach an estimate of the overall quality of the generated answers. 

\begin{center}
$BLEU = min( 1, \frac{output-length}{gold-standard-length}) \prod_1^4 precision_i$
\end{center}

\textbf{METEOR}
(Metric for Evaluation of Translation with Explicit ORdering) is based on the
harmonic mean of precision and recall, whereby recall is weighted higher than
precision. Along with exact word (or phrase) matching it has additional
features, i.e. stemming, paraphrasing and synonymy matching. In contrast to
BLEU, the metric produces good correlation with human judgement at the sentence
or segment level. Differently to the BLEU metric, which penalises the hypothesis, if words were generated with a different inflection, \textbf{chrF3} is a character n-gram metric, which showed very good correlations with human judgements for morphologically rich languages. 

\begin{center}
$chrF\beta = (1 + \beta^2)\frac{chrP*chrR}{\beta^2*chrP+chrR}$
\end{center}

chrP represents the percentage of n-grams in the generated answer which has a counterpart in the gold standard answer, whereby chrR represents the percentage of character n-grams in the gold standard, which are also present in the predicted answer. $\beta$ is a parameter which assigns $\beta-$times more importance to recall than to precision.

%

\begin{table*}[htpb]
\centering
\tiny
\setlength{\tabcolsep}{4pt}
\begin{tabular}{lll}
\toprule
$<$http://dbpedia.org/resource/Aristotle$>$ & $<$http://dbpedia.org/property/name$>$ & "Aristotle"@en . \\
$<$http://dbpedia.org/resource/Aristotle$>$ & $<$http://dbpedia.org/property/birthDate$>$ & "384"$<$http://www.w3.org/2001/XMLSchema\#integer$>$ .\\ 
$<$http://dbpedia.org/resource/Aristotle$>$ & $<$http://dbpedia.org/property/deathDate$>$ & "322"$<$http://www.w3.org/2001/XMLSchema\#integer$>$ .\\ 
$<$http://dbpedia.org/resource/Aristotle$>$ & $<$http://dbpedia.org/property/deathDate$>$ & $<$http://dbpedia.org/resource/Euboea$>$ .\\ 
$<$http://dbpedia.org/resource/Aristotle$>$ & $<$http://dbpedia.org/property/deathDate$>$ & ", Greece"@en .\\ 
$<$http://dbpedia.org/resource/Aristotle$>$ & $<$http://dbpedia.org/property/era$>$ & $<$http://dbpedia.org/resource/Ancient\_philosophy$>$ .\\ 
$<$http://dbpedia.org/resource/Aristotle$>$ & $<$http://dbpedia.org/property/region$>$ & $<$http://dbpedia.org/resource/Western\_philosophy$>$ .\\ 
$<$http://dbpedia.org/resource/Aristotle$>$ & $<$http://dbpedia.org/property/schoolTradition$>$ & $<$http://dbpedia.org/resource/Peripatetic\_school$>$ .\\ 
$<$http://dbpedia.org/resource/Aristotle$>$ & $<$http://dbpedia.org/property/schoolTradition$>$ & $<$http://dbpedia.org/resource/Aristotelianism$>$ .\\ 
$<$http://dbpedia.org/resource/Aristotle$>$ & $<$http://dbpedia.org/property/mainInterests$>$ & $<$http://dbpedia.org/resource/Biology$>$ .\\ 
$<$http://dbpedia.org/resource/Aristotle$>$ & $<$http://dbpedia.org/property/mainInterests$>$ & $<$http://dbpedia.org/resource/Zoology$>$ .\\ 
$<$http://dbpedia.org/resource/Aristotle$>$ & $<$http://dbpedia.org/property/mainInterests$>$ & $<$http://dbpedia.org/resource/Physics$>$ .\\ 
$<$http://dbpedia.org/resource/Aristotle$>$ & $<$http://dbpedia.org/property/mainInterests$>$ & $<$http://dbpedia.org/resource/Metaphysics$>$ .\\ 
$<$http://dbpedia.org/resource/Aristotle$>$ & $<$http://dbpedia.org/property/mainInterests$>$ & $<$http://dbpedia.org/resource/Logic$>$ .\\ 
$<$http://dbpedia.org/resource/Aristotle$>$ & $<$http://dbpedia.org/property/mainInterests$>$ & "Ethics"@en .\\ 
$<$http://dbpedia.org/resource/Aristotle$>$ & $<$http://dbpedia.org/property/mainInterests$>$ & $<$http://dbpedia.org/resource/Rhetoric$>$ .\\ 
$<$http://dbpedia.org/resource/Aristotle$>$ & $<$http://dbpedia.org/property/mainInterests$>$ & "Music"@en .\\ 
$<$http://dbpedia.org/resource/Aristotle$>$ & $<$http://dbpedia.org/property/mainInterests$>$ & "Poetry"@en .\\ 
$<$http://dbpedia.org/resource/Aristotle$>$ & $<$http://dbpedia.org/property/mainInterests$>$ & "Theatre"@en .\\ 
$<$http://dbpedia.org/resource/Aristotle$>$ & $<$http://dbpedia.org/property/mainInterests$>$ & "Politics"@en .\\ 
$<$http://dbpedia.org/resource/Aristotle$>$ & $<$http://dbpedia.org/property/mainInterests$>$ & "Government"@en .\\ 
$<$http://dbpedia.org/resource/Aristotle$>$ & $<$http://dbpedia.org/property/notableIdeas$>$ & $<$http://dbpedia.org/resource/Golden\_mean\_(philosophy)$>$ .\\ 
$<$http://dbpedia.org/resource/Aristotle$>$ & $<$http://dbpedia.org/property/notableIdeas$>$ & $<$http://dbpedia.org/resource/Term\_logic$>$ .\\ 
$<$http://dbpedia.org/resource/Aristotle$>$ & $<$http://dbpedia.org/property/notableIdeas$>$ & $<$http://dbpedia.org/resource/Syllogism$>$ .\\ 
$<$http://dbpedia.org/resource/Aristotle$>$ & $<$http://dbpedia.org/property/notableIdeas$>$ & $<$http://dbpedia.org/resource/Hexis$>$ .\\ 
$<$http://dbpedia.org/resource/Aristotle$>$ & $<$http://dbpedia.org/property/notableIdeas$>$ & $<$http://dbpedia.org/resource/Hylomorphism$>$ .\\ 
$<$http://dbpedia.org/resource/Aristotle$>$ & $<$http://dbpedia.org/property/notableIdeas$>$ & $<$http://dbpedia.org/resource/On\_the\_Soul$>$ .\\ 
$<$http://dbpedia.org/resource/Aristotle$>$ & $<$http://dbpedia.org/property/influences$>$ & $<$http://dbpedia.org/resource/Parmenides$>$ .\\ 
$<$http://dbpedia.org/resource/Aristotle$>$ & $<$http://dbpedia.org/property/influences$>$ & $<$http://dbpedia.org/resource/Socrates$>$ .\\ 
$<$http://dbpedia.org/resource/Aristotle$>$ & $<$http://dbpedia.org/property/influences$>$ & $<$http://dbpedia.org/resource/Plato$>$ .\\ 
$<$http://dbpedia.org/resource/Aristotle$>$ & $<$http://dbpedia.org/property/influences$>$ & $<$http://dbpedia.org/resource/Heraclitus$>$ .\\ 
$<$http://dbpedia.org/resource/Aristotle$>$ & $<$http://dbpedia.org/property/influences$>$ & $<$http://dbpedia.org/resource/Democritus$>$ . \\
\bottomrule
\end{tabular}
\setlength{\tabcolsep}{12.5pt}
\begin{tabular}{ll}
\toprule
Question & Answer \\
\midrule
name of Aristotle & Aristotle \\
birth Date of Aristotle & 384 \\
death Date of Aristotle & 322, Euboea, Greece \\
era of Aristotle & Ancient philosophy \\
religion of Aristotle & Western philosophy \\
school Tradition of Aristotle & Peripatetic school, Aristotelianism \\
main Interests of Aristotle & Biology, Zoology, Physics, Metaphysics, Logic, Ethics, Rhetoric, Music, Poetry, Theatre, Politics, Government \\
notable Ideas of Aristotle & Golden mean, Term logic, Syllogism, Hexis, Hylomorphism, On the Soul \\
influences of Aristotle & Parmenides, Socrates, Plato, Heraclitus, Democritus \\
\bottomrule
\end{tabular}
\caption{Example of DBpedia n-triples \textit{Aristotle} and it's transformation into aligned data format.}
\label{tab:dbp_qa_examples}
\end{table*}

\section{Methodology}
\label{sec:meth}
The training approach to the sequence-to-sequence neural model requires an aligned dataset of questions and answers, which are aligned on a sentence level. To train our QA system on generic knowledge, we extracted this required information from the \textit{DBpedia Infobox Property} repository.

\paragraph{\textbf{DBpedia n-tripple extraction}} We used the \textit{DBpedia Infobox Properties} to extract the data required to train the sequence-to-sequence model. We threat the subject and predicate of each instance in the n-triple repository as a question, whereby the object of the n-triple represents the answer to the question. The top part of Table~\ref{tab:dbp_qa_examples} shows the DBpedia Infobox Properties of the entry \textit{Aristotles}, whereby the bottom part shows the extracted information used to train the neural network. As an example, we extract from \textit{.../resource/Aristotle} (subject of the triple) and \textit{.../property/era} (predicate) the question \textit{era of Aristotle}, and use the object of the n-triple, i.e. \textit{Ancient philosophy}, as the answer of the question.\footnote{We transform all objects into labels, if n-triple object is encoded with an URL link.} Furthermore, we split the properties, which are generated out of several words, at medial capitals (or “CamelCase”), i.e. \textit{birthDate} is transformed into \textit{birth date}. At last, we merge object entries into one single line, if they belong to the same predicate property. As an example, objects of the predicate \textit{main interests}, were grouped together as one entry (answer), i.e. \textit{Biology, Zoology, Physics, Metaphysics, Logic, Ethics, Rhetoric, Music, Poetry, Theatre, Politics, Government}.

\section{Evaluation}
\label{sec:eval}
In this section we report the results of the automatic and a manual evaluation of the answers generated by the neural models.

\subsection{Automatic Evaluation}
The results of the automatic evaluation using the BLEU, METEOR and chrF metric are shown in Table~\ref{tab:auto_eval}. We observer BLEU and Meteor scores, which evaluate the generated answers on based on a n-gram overlap, of 16.48 and 13.97, respectively. BLEU-1 and the chrF score, which perform on a single word or character level, are slightly higher, i.e. 56.0 and 26.84. 

\begin{table}[b]
\centering
\begin{tabular}{cccc}
\toprule
BLEU & BLEU-1 & METEOR & chrF \\
\midrule
16.48 & 56.0 & 13.97 & 26.84 \\
\bottomrule
\end{tabular}
\caption{Automatic evaluation of the neural networks based on the gold standard}
\label{tab:auto_eval}
\end{table}

\begin{table}[t]
\small
\centering
\begin{tabular}{rlr}
\toprule
Avg. METEOR & DBpedia  & Appearances\\
score & Property & in Test Set \\
\midrule
91.15 & next single of & 19 \\
87.09 & basin countries of & 7 \\
85.08 & background of & 34 \\
82.85 & services of & 1 \\
82.00 & direction a of & 5 \\
79.93 & show name of & 20 \\
77.50 & year of & 8 \\
76.32 & official name of & 107 \\
76.26 & genus of & 91 \\
75.07 & media type of & 10 \\
74.33 & conflict of & 8 \\
71.68 & location country of & 6 \\
\dots & \dots & \dots \\
3.07 & nationalteam of & 32 \\
2.88 & composer of & 6 \\ 
2.48 & starring of & 38 \\
1.87 & intercommunality of & 9 \\
1.39 & operating system of & 3 \\
\bottomrule
\end{tabular}
\caption{Statistics on DBpedia Properties, which contributed to the quality of the generated answers based on the METEOR evaluation metric.}
\label{tab:best_meteor_prop}
\end{table}

\begin{table}[t]
\small
\centering
\setlength{\tabcolsep}{2pt}
\begin{tabular}{lr|l}
\toprule
Question & GS & QA Answer \\
\midrule
domain of acidomonas methanolica & bacteria & bacteria \\
capital of korean empire & seoul & seoul \\ 
profession of linda j. lezotte & attorney & lawyer \\
country of chillicothe high school & usa  & united states \\
nationality of jean michel diot & french  & france \\
short description of clark duke & actor  &  football player \\
place of birth of edith roger & vestby,  & vienna, \\
& norway  &  austria \\
\bottomrule
\end{tabular}
\caption{Examples of generated answers compared to the Dbpedia Infobox Property gold standard (GS).}
\label{tab:examples}
\end{table}

\subsection{Manual Evaluation}
\label{subsec:man_eval}

Besides the automatic evaluation of answer generation over the DBpedia Infobox properties, we performed a detailed manual evaluation of the answers produced by the neural networks.

We first evaluated, which answers, based on the DBpedia properties (of the property-subject pair) were correctly generated based on the METEOR score. Table~\ref{tab:best_meteor_prop} illustrates that answers, based on the property \textit{next single of} or \textit{basin countries of} were mostly generated correctly. On the other hand, answers, based on question containing the properties \textit{operating system of} or \textit{intercommunality of} were mostly generated wrongly.

Finally, Table~\ref{tab:examples} presents different examples of answers, based on the Dbpedia Infobox Properties. Although the properties \textit{domain of} or \textit{capital of} appear only once in the evaluation set, they were answered correctly. Furthermore, the neural networks often provide answers, which are synonyms or closely related words to the gold standard answers. As an example, the generated answer \textit{lawyer} is an synonym word of \textit{attorney}, and \textit{usa} can also be referred to as \textit{united states} only. Similarly, a closely related word to \textit{french} is \textit{france}, although this cannot be counted as a correctly generated answer to the question \textit{nationality of jean michel diot}. At last, the proposed approach of using neural networks also demonstrated various wrongly generated answered. As an example, \textit{clark duke} is not an \textit{football player} and \textit{edith roger} was born in \textit{vestby, norway} and not in \textit{vienna , austria} as answered by the proposed neural network approach.

\section{Conclusions}
We presented the work on a QA system trained with a sequence-to-sequence neural model with the DBpedia knowledge and movie dialogues. Although the automatic evaluation shows a low overlap of generated answers compared to the gold standard, a manual inspection of the showed promising outcomes from the experiment. Due to the nature of the training dataset, short answers are preferred, since they are more likely to have a lower log-likelihood score than the longer ones. Nevertheless, we observed several correct answers, which shows the availability of storing the entire DBpedia knowledge into neural networks. Our future work will focus on providing longer answers, as well as focusing on answering more complex questions.

\section*{Acknowledgement}
This publication has emanated from research conducted with the financial support of Science Foundation Ireland (SFI) under Grant Number SFI/12/RC/2289 (Insight).

\newpage

\bibliographystyle{acl_natbib}
\bibliography{refs.bib}

\end{document}